\documentclass{article} 
\usepackage{textcomp}
\usepackage{fontawesome}
\usepackage{iclr2026_conference,times}

\makeatletter
\renewcommand{\ps@headings}{
  \renewcommand{\headrulewidth}{0pt} 
  \renewcommand{\headrule}{\hbox to\headwidth{\hrule height 0.4pt \hfill}} 

  \renewcommand{\@oddhead}{\textit{Preprint}\hfill}
  \renewcommand{\@evenhead}{\@oddhead}
}
\makeatother
\usepackage{amsmath,amsfonts,bm}

\def\eqref#1{equation~\ref{#1}}

\def\1{\bm{1}}

\DeclareMathAlphabet{\mathsfit}{\encodingdefault}{\sfdefault}{m}{sl}
\SetMathAlphabet{\mathsfit}{bold}{\encodingdefault}{\sfdefault}{bx}{n}

\usepackage{hyperref}
\usepackage{url}
\usepackage{inconsolata}
\usepackage{graphicx}
\usepackage{enumitem}
\usepackage{booktabs}
\usepackage{multirow}
\usepackage{wrapfig}        
\usepackage{subcaption}     
\usepackage{makecell}
\usepackage{soul}
\usepackage{colortbl,array,xcolor}
\usepackage{listings}
\usepackage{longtable}

\definecolor{myred}{rgb}{0.7, 0.3, 0.0}
\definecolor{myblue}{HTML}{054488}
\definecolor{mygreen}{HTML}{056b34}
\definecolor{myorange}{HTML}{ff8800}
\definecolor{mypurple}{HTML}{8400ff}
\definecolor{mypink}{HTML}{f7acb9}
\definecolor{deepred}{RGB}{180,0,0}
\definecolor{deepgreen}{RGB}{0,180,0}
\definecolor{myblack}{RGB}{0,0,0}

\lstdefinestyle{python}{
    language=Python,
    basicstyle=\ttfamily\small,
    keywordstyle=\color{blue}\bfseries,
    commentstyle=\color{deepgreen},
    stringstyle=\color{red},
    numberstyle=\tiny\color{gray},
    showstringspaces=false,
    frame=single,
    breaklines=true,
    backgroundcolor=\color{lightgray!15}
}

\title{Toward Effective Tool-Integrated Reasoning via Self-Evolved Preference Learning}

\author{Yifei Chen, Guanting Dong , Zhicheng Dou\thanks{Corresponding author.}\\
Renmin University of China\\
\texttt{\{zhangboguodong, dou\}@ruc.edu.cn}\\
\\
\begin{tabular}{@{}ll@{}}
\faGithub\ GitHub: \href{https://github.com/asilverlight/Tool-Light}{\texttt{\textcolor{cyan}{https://github.com/asilverlight/Tool-Light}}}
\end{tabular}
}

\iclrfinalcopy 
\begin{document}

\maketitle

\begin{abstract}
Tool-Integrated Reasoning (TIR) enables large language models (LLMs) to improve their internal reasoning ability by integrating external tools. However, models employing TIR often display suboptimal behaviors, such as insufficient or excessive tool usage and overthinking after tool calls. The challenge of incentivizing LLMs to perform TIR efficiently and accurately, while stabilizing the reasoning process, remains an open question. In this paper, we start by exploring the impact of tool calls on model reasoning from the perspective of information entropy. Our findings indicate that tool call results lead to a distinct change in the information entropy of subsequent reasoning, with the overall entropy of the reasoning chain varying based on the number of tool calls. Building on these insights, we propose Tool-Light, a framework designed to encourage LLMs to perform TIR efficiently and accurately. Our framework includes dataset construction and multi-stage fine-tuning. For dataset construction, we employ continuous self-evolved sampling using the fine-tuned model, integrating both vanilla sampling and entropy-guided sampling. Besides, we establish strict criteria for selecting positive-negative pairs during sampling. The training process involves a two-stage approach, comprising Supervised Fine-Tuning (SFT) and Self-Evolved Direct Preference Optimization (DPO). Experimental results on 10 datasets demonstrate the effectiveness of Tool-Light, significantly improving the model's efficiency in executing TIR tasks.
\end{abstract}

\section{Introduction}

Recently, large language models (LLMs) have demonstrated strong reasoning abilities after training~\citep{kimiteam,from_system1_2_system2,reasoning_llm_survey,survey_reasoning_foundation_models}. They excel in tasks such as mathematical reasoning~\citep{grpo,rft,mugglemath}, logical puzzles~\citep{logic-rl}, and code generation~\citep{debuglikeahuman,li2025reasoningresourceoptimizingfast}. However, when faced with more challenging tasks (such as deep information retrieval and precise computation~\citep{imitateexplore,o1replicationjourney}), those trained LLMs often struggle when relying solely on their internal reasoning capabilities. To solve those problems, Tool-Integrated Reasoning (TIR)~\citep{art,tora,start,ragen} method has emerged. Compared to general reasoning models, TIR models can autonomously utilize external tools during the reasoning process to compensate for the deficiencies in their internal knowledge or abilities~\citep{research,r1searcher,dotamath,start,searcho1,dparag,arpo}.

While introducing external tools enables models to enhance performance on reasoning tasks, there still exist challenges such as unreasonable tool call or overthinking. For example, during TIR process, the model often exhibit suboptimal tool call patterns, such as excessive or insufficient tool calls~\citep{searchwisely,towardseffectivecode}. Additionally, when low-quality tool call results are provided, models may experience overthinking or even analysis paralysis~\citep{deepresearcher,stopoverthinking,towardsreasoningera}. We collectively refer to them as incorrect tool calls. Recent studies have focused on optimizing tool calls with the help of reinforcement learning (RL)~\citep{towardseffectivecode,feng2025retoolreinforcementlearningstrategic,searchwisely,stepsearch}. However, these studies primarily focus on individual tool, and are hard to generalize to multiple tool call situations. For the TIR task with multiple tool call, those issues may still exist. Some studies have explored reasoning optimization for multiple tool calls~\citep{toolstar,otc}. Nevertheless, existing studies typically focus on the issue of tool-overuse, neglecting both tool underuse and the impact of tool call results on subsequent reasoning processes~\citep{smart}. We consider that these works do not comprehensively address incorrect tool call issues.

At present, many works have analyzed reasoning tasks from the perspective of information entropy~\citep{entropy_mechanism,80_20}. They find that the high-entropy part of the reasoning chain often determines reasoning direction. Inspired by these works, we conduct preliminary experiments and analyze the TIR task's information entropy. We find that after receiving tool call results, the information entropy of the model's subsequent outputs will fluctuate. We also observe that when multiple correct reasoning paths exist, the path with fewer tool calls tends to have a lower overall entropy distribution compared to those with more.
Based on the limitations of existing work and TIR task's entropy characteristics, we attempt to explore the efficiency of the TIR task from two perspectives:

\begin{itemize}[leftmargin=1em]
\item \textbf{From the training perspective:} Can we improve the effectiveness of tool call in TIR tasks from the perspective of post-training (including the algorithm side and the data side)?
\item \textbf{From the inference perspective:} How can we apply the entropy distribution characteristics of the model's TIR process to guide data sampling?
\end{itemize}

In this paper, we introduce Tool-Light, a framework that optimizes the reasoning capability and tool call effectiveness of TIR models from the data construction and algorithmic perspectives. From the data construction perspective, we introduce an innovative entropy-guided sampling method. This approach first generates a main chain, then branches from the highest-entropy positions to create multiple paths. We then apply sampling criteria to derive high-quality positive-negative pairs. To ensure data diversity and balance, we integrate this approach with direct sampling, producing a hybrid method that sustains performance while improving tool-call effectiveness. From the algorithmic perspective, we propose the ``Two-stage TIR Training'' pipeline. First, we perform supervised fine-tuning (SFT) on the original model, and then conduct self-evolved DPO training. In self-evolved DPO stage, we divide the training into two parts: Pre-Aligned DPO Training and Self-Evolved DPO Alignment. The former is employed to enhance the model's reasoning capacity while reducing the redundant tool calls. The latter alternates sampling and training process, enabling the model to learn necessary tool call. At the same time, Self-Evolved DPO Alignment dynamically adjusts the complexity of the training data in accordance with the model's current proficiency, thereby maintaining its original reasoning ability.

To comprehensively evaluate the capabilities of Tool-Light, we use ten challenging reasoning tasks, including knowledge-intensive and mathematical-reasoning tasks. Tool-Light not only ensures stable reasoning performance but also significantly improves the efficiency and accuracy of tool call. 

In summary, our contributions are as follows:

\begin{itemize}[leftmargin=1em]
\item We pioneeringly explore and analyze the TIR paradigm from the perspective of information entropy, demonstrating the connection between TIR effectiveness and entropy change.
\item We propose an innovative entropy-guided sampling strategy, which is combined with a two-stage training method incorporating a self-evolution mechanism, thereby enhancing the effectiveness of the TIR process.
\item Experiment results across 10 challenging reasoning datasets prove the effectiveness of Tool-Light. Further quantitative analyses offer practical guidance for efficient tool-integrated reasoning.
\end{itemize}

\section{Related Work}
\label{related work}

\begin{figure}[t]
  \includegraphics[width=1.\columnwidth]{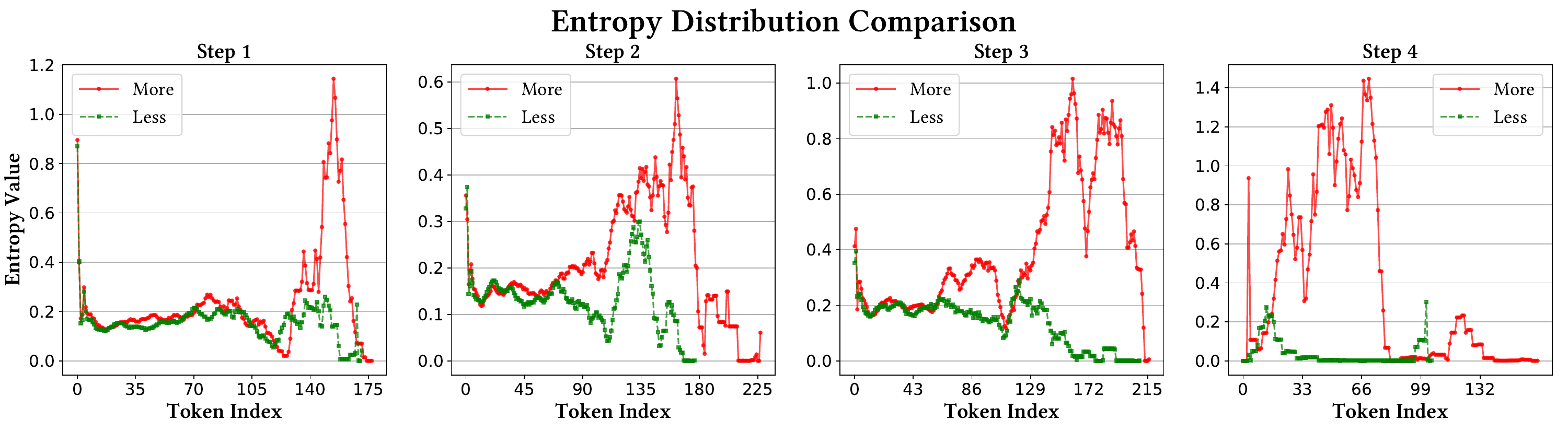}
  \vspace{-1em}
  \caption{The entropy distribution of tool-integrated reasoning inference.}
  \label{fig:preliminary}
\end{figure}
\paragraph{Effective Tool-integrated Reasoning.} Tool-Integrated Reasoning (TIR) refers to guiding models to leverage external tools during reasoning~\citep{codethinkcode}. Teaching models to use tools correctly and rationally is a core challenge in TIR~\citep{CPT,AgentScaler,websailor,websailor-v2,webshaper,webweaver,webresearcher}. Currently, fine-tuning is the main method to guide the model in efficiently completing the TIR task~\citep{chainofagents}. For example, IKEA~\citep{ikea} and SMART~\citep{smart} design training methods based on metacognitive theory~\citep{metacognitive}, focusing on the model's knowledge boundaries. Self-DC framework~\citep{selfdc} controls the model's self-behavior by leveraging internal signals. Additionally, some works efficiently complete the TIR task by optimizing the reinforcement learning training process. For instance, Search Wisely~\citep{searchwisely} and OTC~\citep{otc} carefully designed reward functions during training, while CoRT~\citep{cort} meticulously optimized the training process. Additionally, some work has attempted to expand the TIR domain to explore the field of multimodal reasoning~\citep{thyme,OctoTools,mmsearch-r1,wemath,wemath2}. However, most existing work focuses solely on reducing the excessive tool calls, and does not explore the accuracy of tool call and reasoning process within TIR. We consider that the effectiveness of TIR includes not only reducing redundant tool calls, but also invoking necessary tools. Moreover, it also involves minimizing overthinking during the reasoning process and avoiding analysis paralysis. In this paper, we do not limit our discussion to reducing redundant tool call, but explore the effectiveness of TIR in a broader range of scenarios.
\paragraph{Self-Evolved Methods in LLMs.} Due to the variability and complexity of world knowledge, models need to continuously evolve to better adapt to the real-world tasks~\citep{webdancer,webresearcher,dparag,websailor,webshaper}. The self-evolution strategy refers to a paradigm for dynamically improving model output performance through continuous learning and adaptation~\citep{evolvesurvey}. It can be divided into two methods: parameter-updating and in-context learning. The former involves having the model autonomously generate training data to fine-tune weights, or using the experience from interacting with the environment to drive model training\citep{SCA,SRSI}. The latter achieves model self-evolution through the optimization of prompt words or memory mechanisms\citep{mem0,textgrad}. Referring to existing research, the core of Tool-Light lies in enabling the model to learn from data generated by itself\citep{webthinker}. At the same time, it continuously generates better training data to perform parameter-updating self-evolution.

\section{Preliminaries}\label{preliminary experiments}
\textbf{(1) Problem Definition.} Multi-TIR aims to enable the model to autonomously call multiple tools, thus completing different complex tasks. Specifically, given the input instruction $\mathcal{I}$ and the model $\theta$, the final output answer $y$ of multi-TIR can be expressed in the following form:
\begin{equation}
\small
    y \sim P_{\theta}(y \mid R_N) \cdot \prod_{i = 1}^{N} P_{\theta}(R_i \mid T_{C < i}, R_{< i}, \mathcal{I}),
    \label{eq:modeling}
\end{equation}
where $i$ represents the reasoning step, $N$ represents the total number of reasoning steps, $R_{< i}$, $T_{C < i}$ and $R_{i}$ represent the reasoning content before the $i$-th step, the tool call result before the $i$-th step, and the reasoning content of the $i$-th step, respectively. For the tool $T$, it may contain a variety of different tools in multi-TIR tasks.

\textbf{(2) Pre-Experiment.}
In this section, we explore the relationship between the TIR path and the distribution of information entropy. First, we present the formula for calculating the information entropy distribution:
\begin{equation}
\small
H(i) = - \sum_{j=1}^{N} P(y_{ji} | y_{<i}) log P(y_{ji} | y_{<i}),
\label{eq:token level entropy distribution}
\end{equation}
where $N$ represents the vocabulary length at the $i$-th position, $y_{<i}$ and $y_{ji}$ represent the sequence before position $i$ and the $j$-th token at position $i$, respectively. We use Search-R1~\citep{searchr1} model to perform TIR on multiple QA datasets. For each sample, we rollout ten chains and classify them into two groups based on the tool calls: more tool calls \textbf{(more)} and fewer tool calls \textbf{(less)}. We record the average entropy distribution for each step. Our results can be found in Figure~\ref{fig:preliminary}.

From the results of this experiment, we can draw some conclusions:
\begin{itemize}[leftmargin=1em]
\item When the model receives a tool call result, its output information entropy initially rises, then fluctuates, and finally drops sharply before the next tool call arrives.
\item For the same sample, low-entropy chains tend to involve fewer tool calls, and as reasoning progresses, tool calls' difference between high and low entropy chains becomes increasingly evident.
\end{itemize}
These conclusions provide guidance for our subsequent data sampling.

\textbf{(3) Tool Design} 
Inspired by the existing work, we choose the two most mainstream tools, the code interpreter tool and the search tool, for exploration~\citep{toolstar,otc,torl,searchr1}. The code interpreter tool receives a piece of code, compiles it, and returns the compilation result or error message. The search tool receives a query, and then retrieves the content most relevant to the query from the local knowledge base or web page and returns it.

\section{Method}
\paragraph{Overview.} We propose Tool-Light, a multi-stage training pipeline aiming to improving the effectiveness of model tool calls. As shown in Figures \ref{fig:tree sampling} and \ref{fig:training}, Tool-Light consists of two key components: \textbf{(1) Dataset construction}, which includes carefully designed sampling strategies to screen out training data. \textbf{(2) Two-stage TIR training paradigm}, which trains model successively with SFT and self-evolved DPO training. In the self-evolved DPO training stage, we design pre-aligned DPO training and self-evolved DPO alignment stages to gradually improve the model's capabilities.

\subsection{Dataset Construction}
\label{Entropy-Based Training Set Sampling}
In this section, a detailed introduction to the data construction and sampling strategies are provided.

\begin{wrapfigure}[12]{H}
{0.5\linewidth}
    \centering
    \vspace{-0.5em}
    \includegraphics[width=1\linewidth]{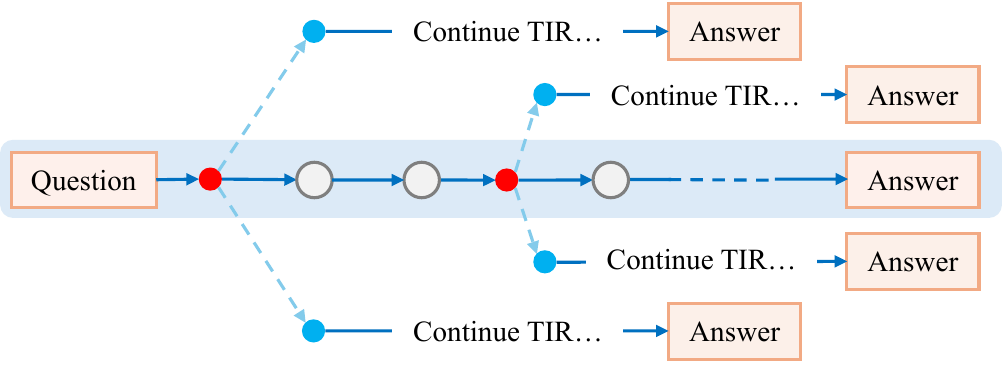}
     \caption{The overall process of entropy-guided sampling. Gray and red positions represent tool calls and the branching positions, respectively.}
    \label{fig:tree sampling}
\end{wrapfigure}

\paragraph{Source Data Construction.} In this section, we first obtain the source data for subsequent sampling. To improve the model's ability to execute TIR tasks, we adopt the SFT data from existing work, defined as $D_{sft}$~\citep{toolstar}. We use $D_{sft}$ to train the instruct model, resulting in the model $M_{sft}$. Then we use the instruct model to perform inference on $D_{sft}$. During this process, we do not offer the model external tools. After that, we only retain the data where the model's inference answers are incorrect, resulting in $D_{source}$. This process can be modeled as:
\begin{equation}
\small
D_{source} = D_{sft}[y_{label} \neq y_{inference}],
\label{eq:make_d_source}
\end{equation}
where $y_{label}$ and $y_{inference}$ are the golden answer and the result obtained from direct inference on that sample, respectively. ${y_{label} \neq y_{inference}}$ means that we only retain data where the model's inference result is incorrect. We consider these datas to be more challenging, and training on them can help better improve the model's reasoning capabilities. 

\paragraph{Sampling Strategy Design.} After obtaining $D_{source}$, we use $M_{sft}$ to perform TIR inference on it. For each question, we sample multiple TIR paths. The final set of all inference paths is denoted as $D_{dpo}$. For this step, we carefully design two sampling strategies, and mix the data from both strategies to construct high-quality training data. The specific strategies are described as follows:

\textbf{(1) Vanilla TIR Sampling.} We guide the model to generate multiple paths for each sample. This helps us construct positive-negative pairs. This process can be expressed as follows:
\begin{equation}
\small
D_{dpo}^1=\{y|y=M_{sft}[I(q)], q \sim D_{source}\},
\label{eq:make_d_dpo_1}
\end{equation}
where $D_{dpo}^1$ represents the sampling results generated under vanilla sampling, $I$, $q$ and $y$ represent the prompt used, each question in $D_{source}$, and the output result from $M_{sft}$, respectively.

\textbf{(2) Entropy-Guided Sampling.} Vanilla sampling strategy incurs high inference costs and inference uncertainty. 
In this section, as shown in Figure~\ref{fig:tree sampling}, we propose an entropy-guided sampling strategy.
\begin{itemize}[leftmargin=1em]
\item \textbf{Entropy Distribution Calculation.} We begin by generating a primary reasoning chain, denoted as $C_{main}$. Each segment of the chain is considered a step. According to existing research~\citep{beginning}, the initial inference step has a significant impact on the subsequent outputs. Therefore, 
for each step, we compute entropy distribution at each token position in the first 10, 20, 30, 40, and 50 tokens. The average information entropy distribution for a subsequence is as follows:
\begin{figure*}[t]
    \centering
    \includegraphics[width=1\linewidth]{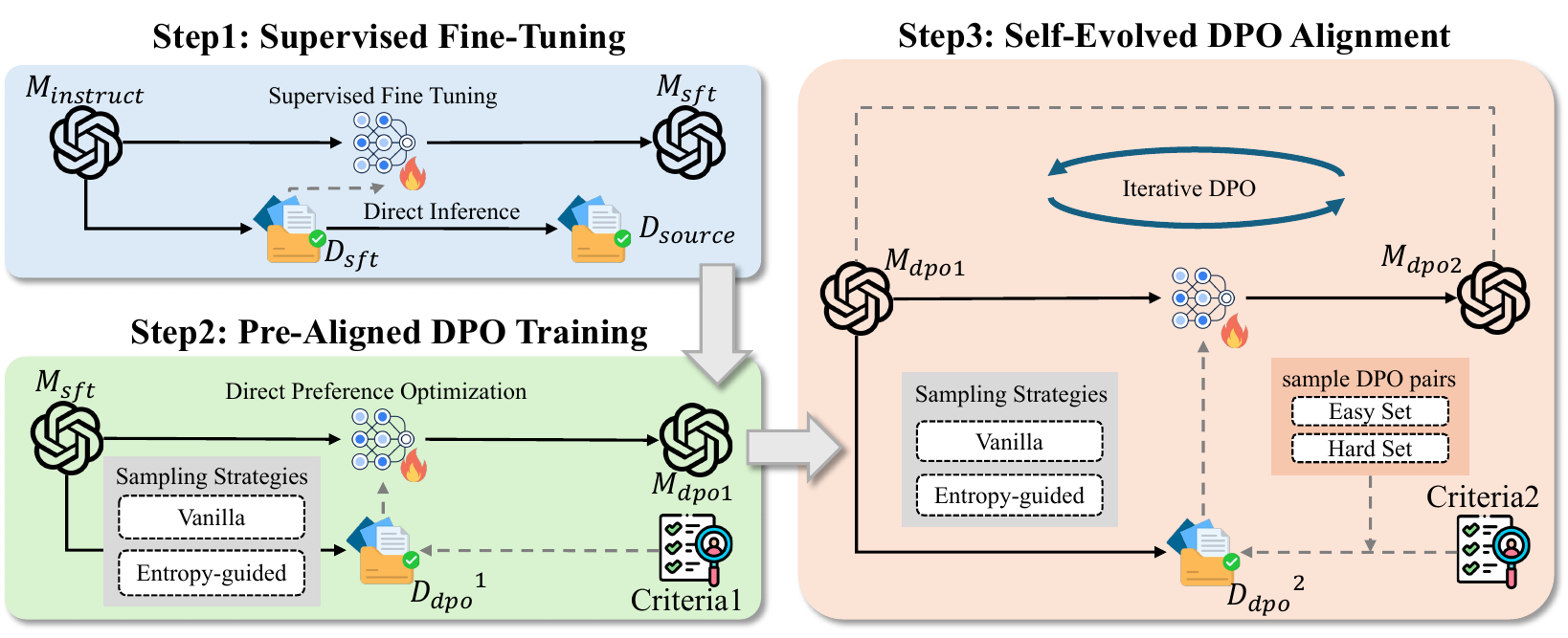}
    \vspace{-0.5em}
    \caption{The overall of Tool-Light's training pipeline. Among them, the Self-Evolved DPO Alignment stage will conduct multiple rounds of training.}
    \label{fig:training}
\end{figure*}
\begin{equation}
\small
H_{avg}(i) = \frac{1}{i} \sum_{j=1}^{i} H(j),
\label{eq:sequence entropy distribution}
\end{equation}
where $H_{avg}(i)$ represents top-$i$ tokens' average entropy distribution. We retain the maximum $H_{avg}(i)$ and its corresponding sequence length $i$ at each step. 
\item \textbf{Branch Sampling Execution.} We pick the top-$k$ steps with the highest entropy, along with their sampling positions. At each position, we guide the model to continue generating several branches. Since those subsequences have higher information entropy, based on previous research~\citep{80_20,entropy_mechanism,cheng2025reasoningexplorationentropyperspective}, it is more likely to produce branch outputs with higher diversity. This process can be modeled as follows:
\begin{equation}
\small
D_{dpo}^2=\{y|y_{>i}=M_{sft}[I(q)\oplus y_{<i}], q \sim D_{source}\},
\label{eq:branch sample}
\end{equation}
where $D_{dpo}^2$ and $y_{>i}$ represent the dataset guided by entropy sampling and the subsequent sampling output, respectively. $I(q)\oplus y_{<i}$ represents concatenating the prompt and the existing sequence together as the input. $y$ represents concatenating $y_{>i}$ and $y_{<i}$ together. Assuming $m$ rollout times and an average sequence length of $n$, this method, in the ideal scenario, reduces computational complexity from $O(mn)$ to $O(n \log m)$. We collect all the main chains and branch chains, and construct the set of outputs based on tree-structured sampling. 
\end{itemize}

Finally, $D_{dpo}^1$ and $D_{dpo}^2$ are mixed in a certain ratio to obtain the final dataset ${D_{dpo}}^1$ for the positive-negative pairs construction. 


\subsection{Two-Stage TIR Training Paradigm}
\label{sec:training strategy}

Based on existing research~\citep{webthinker,toolstar,r1searcher}, we propose a two-stage self-evolved training pipeline to gradually boost the effectiveness and stability of the model's TIR process. The specific pipeline is shown in Figure~\ref{fig:training}.

\paragraph{Supervised Fine-Tuning.}
First, we obtain the SFT model $M_{sft}$ with the same setup as in ToolStar~\citep{toolstar}. The SFT's loss function is: $\mathcal{L}_{\text{SFT}}(\theta) = - \sum_{(x, y) \in D} \log P_{\theta}(y|x)$. As shown in the first step of Figure~\ref{fig:training}, this step aims to help the model quickly acquire the ability to complete TIR tasks. 

\paragraph{Self-Evolved DPO.}
Inspired by the existing work~\citep{webthinker,self-evolved1,self-evolution-llm}, we further fine-tune $M_{sft}$ through a self-evolved DPO strategy. The DPO's loss function is as follows:

\begin{equation}
    \mathcal{L}_{\text{DPO}}(\pi_{\theta}; \pi_{\text{ref}}) = - \mathbb{E}_{(x, y_w, y_l) \sim D} \left[ \log \sigma \left( \beta \log \frac{\pi_{\theta}(y_w|x)}{\pi_{\text{ref}}(y_w|x)} - \beta \log \frac{\pi_{\theta}(y_l|x)}{\pi_{\text{ref}}(y_l|x)} \right) \right],
    \label{eq:dpo}
\end{equation}
where $\pi_{\text{ref}}$ is the original strategy model, and $\beta$ is the hyperparameter and $\sigma$ is the sigmoid function. The primary objective of DPO is to maximize the relative probability of the model generating $y_w$ compared to $y_l$.
This step aims to endow the model with the capability to finish TIR tasks accurately and efficiently. First we establish a specialized data sampling criterion $Cri_1$. It is used to obtain training data ${D_{dpo}}^1$ from $M_{sft}$ using both vanilla and entropy-guided strategies. We then train $M_{sft}$ on ${D_{dpo}}^1$ to yield $M_{dpo1}$. Subsequently, a new criterion $Cri_2$ is defined, guiding $M_{dpo1}$ to sample the next iteration of training data ${D_{dpo}}^2$. We train $M_{dpo1}$ on this dataset and get $M_{dpo2}$. After that we update $M_{dpo1} \leftarrow M_{dpo2}$ and iteratively repeat the sampling and training steps until performance converges. Details on these two criteria follow in the subsequent sections.

\textbf{\textit{(1) Pre-Aligned DPO Training.}}
In each sample, we first categorize the correct or incorrect trajectories based on F1 score. Only trajectories with 1 F1 score are considered correct, and only trajectories with 0 F1 score are considered incorrect. Then, we divide the samples into a hard set and an easy set.
After that, referring to the conclusions in Section~\ref{preliminary experiments}, we design criteria for sampling positive and negative pairs in entropy-guided strategy:
\begin{itemize}[leftmargin=1em]
\item \textbf{Hard set:} Samples with $\le 50\%$ correct trajectories.
\item \textbf{Easy Set:} Samples with $\ge 50\%$ correct trajectories.
\item \textbf{Positive Example:} Correct trajectory with the minimal number of tool calls and lowest entropy. If no correct trajectory exists, the corresponding SFT trajectory from $D_{source}$ is used.
\item \textbf{Negative Example:} An incorrect trajectory with more tool call than the positive example.
\end{itemize}

For vanilla sampling criteria, we choose samples with $\le 40\%$ correct trajectories as hard set, and those with $\ge 70\%$ correct trajectories as easy set. For each set, the shortest correct trajectory is the positive example, and the incorrect trajectory longer than the positive one is the negative example.

We set the hard set and easy set's data ratio as 2:1. The goal of training in this round is to teach the model to reduce unnecessary tool call while avoiding excessive reasoning.

\textbf{\textit{(2) Self-Evolved DPO Alignment.}}
\label{self-evolved dpo loop}
We use $M_{dpo1}$ to resample from ${D_{dpo}}^1$. At the same time, we introduce second criteria to filter out ${D_{dpo}}^2$. In $Cri_2$, samples whose number of correct trajectories is less than half of the incorrect ones are classified into the hard set, while the criteria for the easy set remain unchanged. For each sample generated by $M_{dpo1}$, if it is classified into the easy set, it means that after pre-aligned DPO training, $M_{dpo1}$ has mastered the knowledge to solve this problem. If it is classified into the hard set, it means that this problem is still difficult for $M_{dpo1}$. For these two different situations, we have also designed the sampling criteria for positive and negative examples:
\begin{itemize}[leftmargin=1em]
\item \textbf{Negative Example in Easy Set:} Incorrect trajectory with the most tool calls.
\item \textbf{Positive Example in Easy Set:} Correct trajectory with fewer tool calls and lowest entropy.
\item \textbf{Positive Example in Hard Set:} Correct trajectory with the longest reasoning chain.
\item \textbf{Negative Example in Hard Set:} Incorrect trajectory with the shortest reasoning chain.
\end{itemize}
Based on preliminary experiments, low-entropy outputs' tool calls are often fewer. Therefore, we also incorporate the consideration of low entropy into positive examples' selection. We use the sampled ${D_{dpo}}^2$ to continue training $M_{dpo1}$, and continuously sample with the trained model. Self-evolved DPO alignment will continue for several loops. This step is intended to equip the model with the ability to make necessary tool calls while maintaining its original efficient TIR capability. After several rounds of training, we obtain the final model $M_{dpo2}$.

\section{Experiments}

\subsection{Experimental Setup.}

\paragraph{Dataset.} We carefully select 10 datasets and categorize them into two types: \textbf{(1) mathematical-reasoning tasks}, including AIME24, AIME25~\footnote{\url{https://huggingface.co/datasets/AI-MO/aimo-validation-aime}}, AMC23~\footnote{\url{https://huggingface.co/datasets/zwhe99/amc23}}, MATH~\citep{MATH}, MATH500~\citep{math500}, and GSM8K~\citep{cobbe2021gsm8k}, and \textbf{(2) knowledge-intensive tasks}, including HotpotQA~\citep{hotpotqa}, 2WikiMultiHopQA~\citep{2wiki}, MuSiQue~\citep{musique}, and Bamboogle~\citep{bamboogle}.
\paragraph{Metrics.}
For mathematical-reasoning tasks, we adopt the LLM-as-Judge approach~\citep{llm-as-judge}, utilizing the Qwen2.5-72B-Instruct model to evaluate answer correctness. For knowledge-intensive tasks, we use F1 score to evaluate answers. Additionally, we introduce two metrics: \textbf{Efficiency} and \textbf{Necessity}. \textbf{(1) Efficiency} is defined as $Effi = \frac{1}{n} \sum_{i=1}^{n} \frac{M_i}{T_i}$, where $n$ is the total number of samples, $M_i$ the performance on the $i$-th sample, and $T_i$ the number of tool calls. It captures the model’s tendency to overuse tools. \textbf{(2) Necessity} is defined as $Nece = \mathcal{M}\left( \frac{1}{n} \sum_{i=1}^{n} (N_{in}^i - N_{co}^i) \right)$, where $N_{in}^i$ is the count of methods for the $i$-th sample that make more tool calls than the current method but yield incorrect answers, $N_{co}^i$ is the count of methods with fewer tool calls yet correct answers, and $\mathcal{M}$ denotes Min-Max Scaling. It reflects the model’s tendency to underuse tools.

\paragraph{Baselines.}We classify baseline methods into two types: \textbf{(1) Single-Tool-Integrated Reasoning}, including Search-o1~\citep{searcho1}, Search-R1~\citep{searchr1}, DotaMath~\citep{dotamath}, ToRL~\citep{torl}, and ReTool~\citep{retool}, and \textbf{(2) Multi-Tool-Integrated Reasoning}, including prompt engineering, ReCall~\citep{research}, and ToolStar~\citep{toolstar} methods.\footnote{More implementation details can be found in Appendix~\ref{app:more details of main results}.}

\subsection{Main Result}
Our main results are presented in Table \ref{tab:main_table}, and more details can be found in Appendix~\ref{app:complete res for table1}. We can draw the following insights:


\begin{table*}[!t]
\centering
\caption{Results on 10 reasoning tasks, with top two results highlighted in \textbf{bold} and \underline{underlined}. Unless noted, Qwen2.5-7B-Instruct is used as the backbone. For ToRL and ReTool, their RL models generate training data inferences for SFT, which serves as the baseline. Abbreviation: 2Wiki. (2WikiMultiHopQA). HQA. (HotpotQA). MSQ. (MuSiQue). Bamb. (Bamboogle). 
}
\setlength\tabcolsep{2.3pt}
\renewcommand{\arraystretch}{1} 
\fontsize{8pt}{10.5pt}\selectfont
\begin{tabular}{p{2.6cm}cccccccccccc}
\toprule
\multirow{2}[2]{*}{\textbf{Method}} & \multicolumn{6}{c}{\textbf{Mathematical-Reasoning Tasks}} & \multicolumn{4}{c}{\textbf{Knowledge-Intensive Tasks}} & \multirow{2}[2]{*}{\textbf{Avg.}} \\
\cmidrule(lr){2-7} \cmidrule(lr){8-11}
 & AIME24 & AIME25 & AMC23 & MATH & MATH500 & GSM8K & HQA & 2Wiki. & MSQ & Bamb & \\
\midrule
\multicolumn{12}{l}{\textit{\textbf{Direct Inference}}} \\
Qwen2.5-7B-Instruct & 0.0 & 6.7 & 30.0 & 68.6 & 57.2 & 71.4 & 26.1 & 25.6 & 7.9 & 36.5 & 33.0\\
Llama3.1-8B-Instruct & 0.0 & 3.3 & 15.0 & 52.8 & 33.4 & 75.0 & 16.2 & 13.7 & 7.4 & 23.2 & 24.0\\
\midrule
\multicolumn{12}{l}{\textit{\textbf{Single-TIR Methods}}} \\
Search-o1 & 6.7 & 10.0 & 37.5 & 73.6 & 61.8 & 80.2 & 41.1 & 35.4 & 13.2 & 39.8 & 39.9\\
Search-R1 & 16.7 & 6.7 & 45.0 & 81.2 & 63.8 & 82.4 & 48.7 & 40.0 & \underline{24.1} & 47.4 & 45.6 \\
DotaMath & 16.7 & 10.0 & 50.0 & 74.6 & 62.2 & 82.6 & 26.2 & 21.7 & 6.5 & 28.6 & 37.9\\
ToRL & \underline{30.0} & \underline{26.7} & \textbf{67.5} & \underline{87.0} & \textbf{80.2} & 89.2 & 41.3 & 35.4 & 9.5 & 36.9 & 50.4\\
ReTool & 23.3 & \textbf{30.0} & 62.5 & 84.8 & 78.4 & 86.2 & 31.5 & 29.0 & 11.1 & 35.8 & 47.3\\
\midrule
\multicolumn{12}{l}{\textit{\textbf{Multi-TIR Methods}}} \\
Prompting-Based & 6.7 & 13.3 & 47.5 & 73.8 & 62.2 & 69.4 & 21.1 & 23.8 & 9.9 & 25.5 & 35.3\\
ReCall & 3.3 & 6.7 & 27.5 & 73.2 & 54.6 & 79.8 & 51.9 & 54.0 & \textbf{25.0} & 55.5 & 43.2 \\ 
Tool-Star &  \underline{30.0} & \underline{26.7} & \underline{65.0} & 85.6 & 77.2 & \underline{89.4} & \underline{54.7} & \underline{55.7} & 22.8 & \textbf{58.8} & \underline{56.6}\\
\midrule
\multicolumn{12}{l}{\textit{\textbf{Ours}}} \\
Tool-Light (Llama) & 10.0 & 6.7 & 30.0 & 59.4 & 56.8 & 76.6 & 41.3 & 33.5 & 12.2 & 41.3 & 36.8\\
\rowcolor[RGB]{236,244,252}Tool-Light (Qwen) & \textbf{33.3} & 23.3 & \textbf{67.5} & \textbf{87.4} & \underline{79.0} & \textbf{92.0} & \textbf{57.7} & \textbf{56.1} & \textbf{25.0} & \underline{58.7} & \textbf{58.0}\\
\bottomrule
\vspace{-4em}
\end{tabular}

\label{tab:main_table}
\end{table*}
\textbf{(1) Proper use of external tools can be of great help to model reasoning.} For example, when external tools are introduced, the prompting-based method on the Qwen model has significant advantages over the direct inference method in mathematical reasoning tasks. However, for knowledge-intensive tasks, the effect of introducing external tools is reduced (only improved on MuSiQue dataset). This also indicates that untrained models cannot use external tools well.

\textbf{(2) While training enhances the model's TIR capabilities, single-tool training hinders the development of generalization abilities.} Compared to using prompt engineering methods, fine-tuning can effectively teach the model to use tools. For example, Search-R1 outperforms Search-o1 by 5.7 on average, while Tool-Star outperforms prompting-based methods by 21.3 on average. It is worth noting that models trained for single-TIR tend to have poor generalization performance. For example, Search-R1 performs well on knowledge-intensive tasks but poorly on mathematical-reasoning tasks, while ToRL shows the opposite pattern. The Tool-Star and Tool-Light frameworks perform well on both types of tasks, highlighting the generalization advantages of multi-TIR training.

\textbf{(3) Multi-round DPO training allows for further enhancement of the model's performance.}
Compared with the simple SFT method, Tool-Light framework additionally introduces two operations: Pre-Aligned DPO Training and Self-Evolved DPO Alignment. This leads to an overall improvement in the model's performance. For example, in mathematical-reasoning tasks, Tool-Light achieves optimal performance on four datasets, and ranks among the top two on all datasets for Knowledge-Intensive tasks. Notably, by solely adopting the DPO method, we outperform most baselines that are trained using GRPO when it comes to average performance. This fully demonstrates the effectiveness of our training pipeline design.


\begin{figure}[t]
  \includegraphics[width=1.\columnwidth]{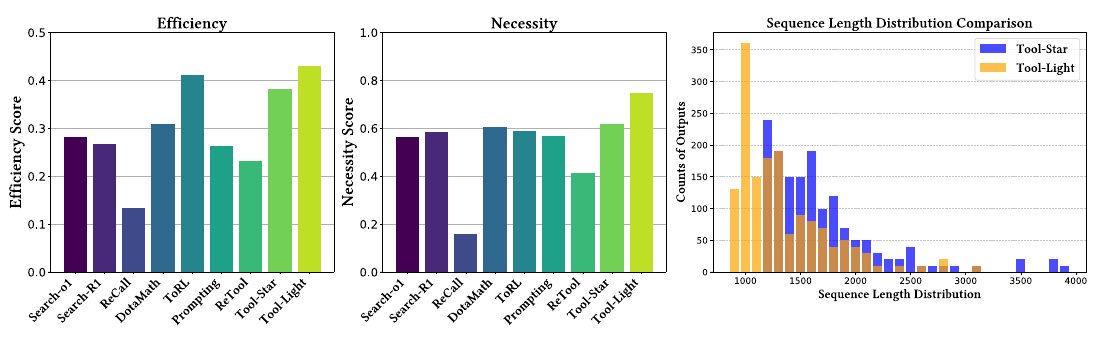}
  \vspace{-2em}
  \caption{The Efficiency, Necessity and Sequence Length differences between Tool Light and baseline methods.}
  \label{fig:main_column}
\end{figure}


\begin{figure}[t]
  \includegraphics[width=\columnwidth]{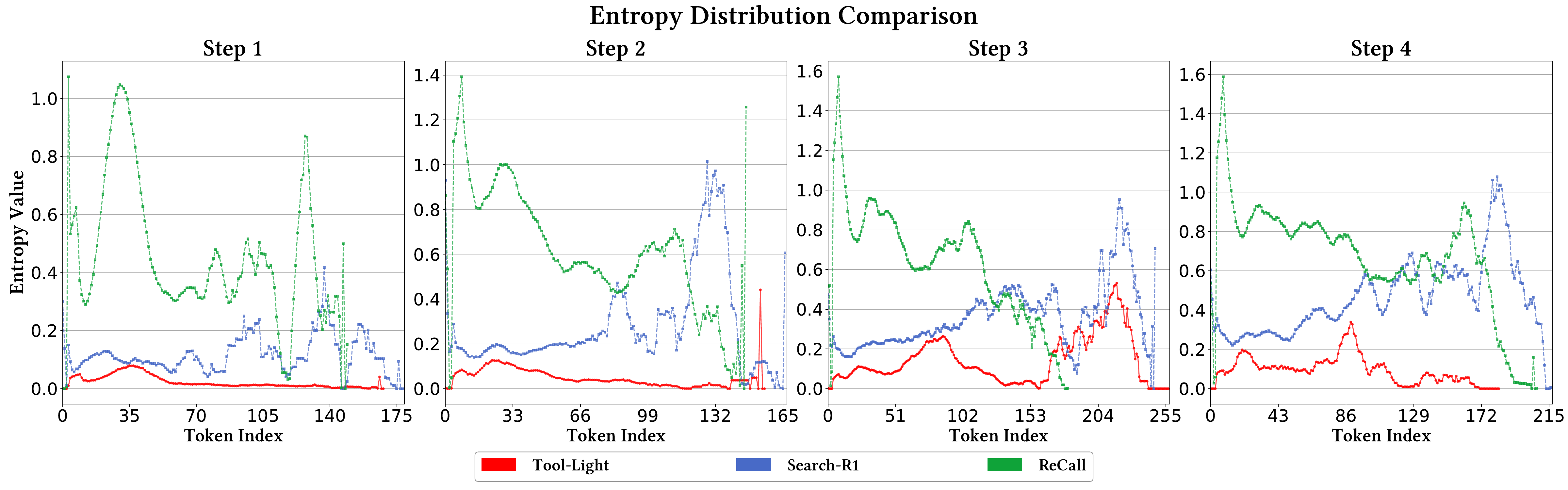}
  \vspace{-1.5em}
  \caption{Output sequences' entropy distribution under different methods.}
  \label{fig:entropy_ablation}
\end{figure}

\subsection{Quantitative Analysis}

\paragraph{Tool-Use Effectiveness Analysis}
In this section, we analyze the differences in TIR effectiveness between Tool-Light and other baseline methods. As shown in Figure~\ref{fig:main_column}, Tool-Light achieves the best performance in Efficiency and Necessity metrics, which indicates the superiority of our designed training process. 
Regarding the output sequence length metric, Tool-Light has a shorter length distribution than Tool-Star, and outperforms Tool-Star in both result correctness and reasoning effectiveness. It shows that our training method can effectively reduce the phenomenon of overthinking. Detailed cases are list in Appendix~\ref{app:case study}.

\paragraph{Entropy Distribution Analysis}
As described in Section~\ref{preliminary experiments}, in this section, we explore the impact of introducing the entropy-guided sampling strategy on final output results' information entropy distribution. The results are shown in Figure \ref{fig:entropy_ablation}. The existing TIR models tend to produce output sequences with relatively high entropy. In contrast, Tool-Light results in lower entropy distributions for its output sequences. This might be due to the consideration of information entropy during sampling. Our training method introduces the learning of low-entropy path outputs. Since high-entropy tokens can sometimes trigger excessive reasoning in the model, our method effectively mitigates the phenomenon of overthinking. This also aligns with the results in Figure \ref{fig:main_column}. \footnote{We also explore the impact of different data ratios on the final training results during sampling. The experimental results and analyses are provided in Appendix~\ref{app:data ratios}.}




\begin{wraptable}{r}{0.55\textwidth}
\centering
\small
\vspace{-1.5em}
\caption{Ablation experiment results for various aspects: \textit{1/1 strategy ratio} indicates a 1:1 data ratio for two sampling strategies, \textit{p-r.} and \textit{n-r.} denote random positive and negative example selection. }
\label{tab:ablation study}
\setlength\tabcolsep{0.5pt} 
\renewcommand{\arraystretch}{1}
\begin{tabular}{p{2.5cm}ccc}
\toprule
\textbf{Method} & \textbf{Performance} & \textbf{Efficiency} & \textbf{Necessity} \\ 
\midrule
Tool-Light (\textit{2 loop}) & 58.0 & 0.44 & 0.75 \\
\midrule
\multicolumn{4}{l}{\textit{\textbf{Ablation for self-evolved Loops}}} \\
\textit{w. 1 loop} & $57.9_{\textcolor{deepred}{\,\text{(-0.1)}}}$ & $0.42_{\textcolor{deepred}{\,\text{(-0.02)}}}$ & $0.71_{\textcolor{deepred}{\,\text{(-0.04)}}}$\\
\textit{w. 3 loop} & $56.1_{\textcolor{deepred}{\,\text{(-1.9)}}}$ & $0.39_{\textcolor{deepred}{\,\text{(-0.05)}}}$ & $0.73_{\textcolor{deepred}{\,\text{(-0.02)}}}$ \\
\textit{w. 4 loop} & $56.4_{\textcolor{deepred}{\,\text{(-1.6)}}}$ & $0.37_{\textcolor{deepred}{\,\text{(-0.07)}}}$ & $0.71_{\textcolor{deepred}{\,\text{(-0.04)}}}$ \\
\textit{w. 5 loop} & $54.1_{\textcolor{deepred}{\,\text{(-3.9)}}}$ & $0.36_{\textcolor{deepred}{\,\text{(-0.08)}}}$ & $0.72_{\textcolor{deepred}{\,\text{(-0.03)}}}$ \\
\midrule
\multicolumn{4}{l}{\textit{\textbf{Ablation for Sampling Criteria}}} \\
\textit{w. 1/1 data ratio} & $56.9_{\textcolor{deepred}{\,\text{(-1.1)}}}$ & 0.44 & $0.76_{\textcolor{deepgreen}{\,\text{(+0.01)}}}$\\
\textit{w. p-r.} & $53.6_{\textcolor{deepred}{\,\text{(-4.4)}}}$ & $0.42_{\textcolor{deepred}{\,\text{(-0.02)}}}$ & $0.63_{\textcolor{deepred}{\,\text{(-0.12)}}}$ \\
\textit{w. n-r.} & $53.9_{\textcolor{deepred}{\,\text{(-4.1)}}}$ & $0.41_{\textcolor{deepred}{\,\text{(-0.03)}}}$ & $0.74_{\textcolor{deepred}{\,\text{(-0.01)}}}$ \\
\bottomrule
\vspace{-10em}
\end{tabular}
\end{wraptable}

\paragraph{Impact of DPO Loop Number}
In this section, we investigate how the number of training loops in the self-evolved DPO alignment stage influences the model's performance. For this stage, we increase the loop count from 1 to 5 and record the Performance, Efficiency, and Necessity metrics. The results are shown in Table \ref{tab:ablation study}'s 1-5 rows. As the number of training loop increases, all metrics reach optimal values in the second training round and then decline. This may be because the self-evolved method is initially able to obtain sufficient positive-negative pairs. However, as the evolution proceeds, there are fewer sample pairs beneficial for training, and the model may overfit to the distribution of the training set, leading to lower performance on the test set.


\paragraph{Impact of Sampling Criteria}
In this section, we explore the impact of sampling criteria on the training effect. We modify the original criteria from the following aspects: the data ratio of two strategies is 1:1, positive examples are randomly selected, and negative examples are randomly selected. The results are shown in rows 6 - 8 of Table \ref{tab:ablation study}. We can find that the different data ratio has the least impact on the training effect, while the change in the positive-negative example selection criteria has a greater impact on the results. This indicates that for the DPO training in Tool-Light, it is very important to fully distinguish between positive and negative examples.

\section{Conclusion}

In this paper, we propose Tool-Light, a framework empowering models to efficiently and accurately complete TIR tasks. We first analyze the TIR process from the perspective of information entropy. Based on this analysis, we introduce Tool-Light, a framework improving model's TIR capabilities from both data construction and training aspects. Within this framework, we design a new entropy-guided sampling strategy. Based on this method, we construct a two-stage TIR training paradigm. The trained model achieves excellent results in terms of overall performance, tool calls effectiveness, and simplicity of the inference process. We confirm that our work can provide more assistance for future research on TIR.




\bibliography{iclr2026_conference}
\bibliographystyle{iclr2026_conference}

\clearpage
\appendix


\section{More Details of Main Results}
\label{app:more details of main results}
\paragraph{Implementation Details.}
We use LoRA training method~\citep{lora}. During training, the LoRA rank is set to 8, the epoch is set to 3, the learning rate is 7e-6, and the batch size is 8. For the self-evolved DPO alignment, we train for two rounds. During testing, we set the temperature to 1 and limit the maximum call count for each tool to 4. For knowledge-intensive tasks, we use E5-base-v2~\citep{e5} with Wikipedia retrieval based on FlashRAG's settings~\citep{flashrag}. For mathematical-reasoning tasks, we use Bing Web Search for retrieval. To avoid overly lengthy during web search, we introduce Web Browser mechanism following Tool-Star.
\paragraph{Full Results at Tool-Light's Different Stages}
We present the full results of Qwen2.5-7B-Instruct model at Tool-Light framework's different stages. The results can be found in Table~\ref{app:complete res for table1}. From these results, we can draw the following insights: (1) As each training phase of Tool-Light progresses, the performance of the model gradually improves. For example, the performance of the model has significantly improved after SFT. After multiple rounds of DPO training, its performance further increases. (2) As the self-evolved DPO alignment loops proceed, the model's performance gradually converges. The performance of the model with the self-evolved DPO alignment is better than with only pre-aligned DPO training. However, with the increase of self-evolution rounds, the model's performance improves slowly. This may be because it is difficult to further generate samples beneficial for model training, resulting in only a small improvement in the model's performance after training.

\begin{table*}[!t]
\centering
\setlength\tabcolsep{2pt}
\renewcommand{\arraystretch}{1} 
\fontsize{8pt}{10.5pt}\selectfont
\begin{tabular}{p{3cm}cccccccccccc}
\toprule
\multirow{2}[2]{*}{\textbf{Method}} & \multicolumn{6}{c}{\textbf{Mathematical-Reasoning Tasks}} & \multicolumn{4}{c}{\textbf{Knowledge-Intensive Tasks}} & \multirow{2}[2]{*}{\textbf{Avg.}} \\
\cmidrule(lr){2-7} \cmidrule(lr){8-11}
 & AIME24 & AIME25 & AMC23 & MATH & MATH500 & GSM8K & HQA & 2Wiki. & MSQ & Bamb & \\
\midrule
Prompting-Based & 6.7 & 13.3 & 47.5 & 73.8 & 62.2 & 69.4 & 21.1 & 23.8 & 9.9 & 25.5 & 35.3\\
Tool-Light (w. SFT) &  \underline{30.0} & \textbf{26.7} & \underline{65.0} & 85.6 & 77.2 & 89.4 & 54.7 & 55.7 & 22.8 & 58.8 & 56.6\\
Tool-Light (w. PA.) & \underline{30.0} & \textbf{26.7} & \textbf{67.5} & 85.8 & \textbf{80.4} & 90.8 & 55.5 & \textbf{56.5} & 24.3 & 56.0 & 57.4\\
Tool-Light (w. SE.)\\
\rowcolor[RGB]{236,244,252}\quad\textit{w. SE. 1 Loop} & \underline{30.0} & \textbf{26.7} & \textbf{67.5} & \underline{86.4} & 78.2 & \textbf{92.2} & \underline{56.7} & 55.6 & \textbf{26.4} & \textbf{59.5} & \underline{57.9}\\
\rowcolor[RGB]{236,244,252}\quad\textit{w. SE. 2 Loop} & \textbf{33.3} & \underline{23.3} & \textbf{67.5} & \textbf{87.4} & \underline{79.0} & \underline{92.0} & \textbf{57.7} & \underline{56.1} & \underline{25.0} & \underline{58.7} & \textbf{58.0}\\
\bottomrule
\vspace{-3em}
\end{tabular}
\caption{Full results of Tool-Light on 10 reasoning tasks, with best two results highlighted in \textbf{bold} and \underline{underline}. We use Qwen2.5-7B-Instruct as backbone model. Abbreviation: 2Wiki. (2WikiMultiHopQA). HQA. (HotpotQA). MSQ. (MuSiQue). Bamb. (Bamboogle). SE. (only conducting pre-alignment DPO training after SFT). PA. (conducting pre-alignment training and several self-evolved DPO alignment loops after SFT).
}
\label{app:complete res for table1}
\end{table*}

\section{Impact of Different Data Ratios}
\label{app:data ratios}
\begin{figure}[t]
  \includegraphics[width=\columnwidth]{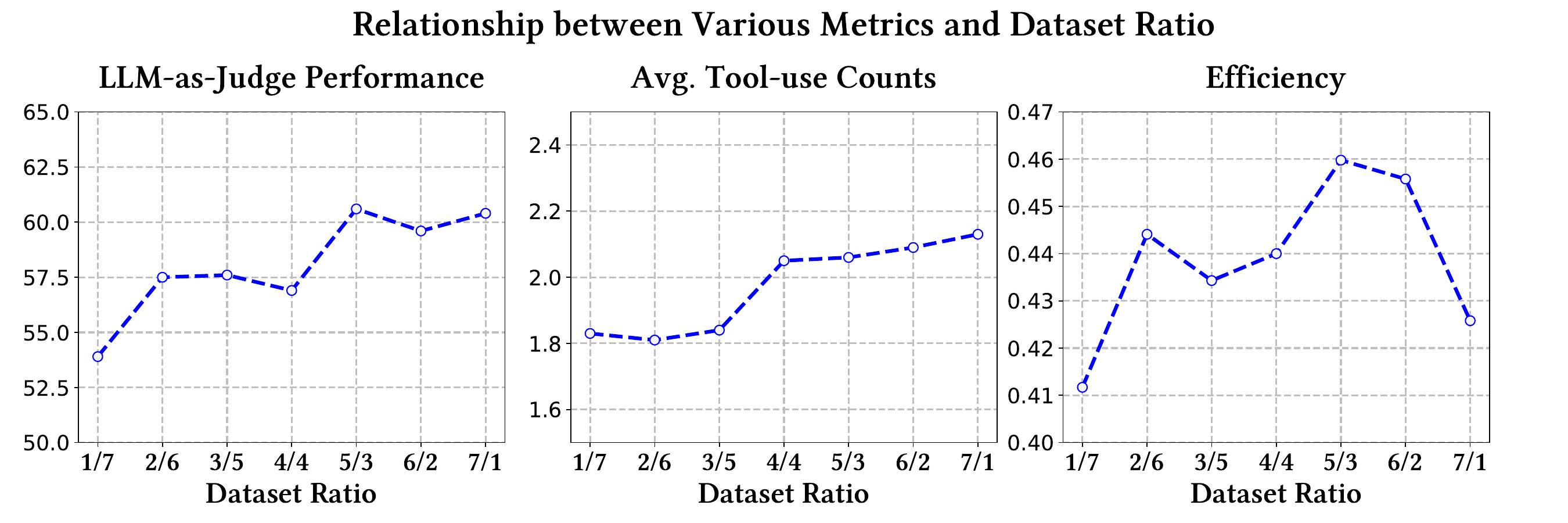}
  \vspace{-1.5em}
  \caption{The variation of metrics with changes in the data ratio.}
  \label{fig:data ratio}
\end{figure}

In this section, we explore different data ratios' impact of two sampling methods on training performance. We sample data using the vanilla method and the entropy-guided method separately. Then, we mix the data in ratios ranging from 1:7 to 7:1. Meanwhile, we ensure that other settings remain unchanged. The model is then retrained in the first round of DPO, and the results are shown in Figure \ref{fig:data ratio}.

Results show that the model's performance in the TIR task is strongly related to the mixing ratio of the two data types. As the proportion of data with the vanilla method increases, both the model's performance and the number of tool calls improve. This is likely because training on this type of data makes the model more inclined to use tools to solve problems. However, the Efficiency metric does not continuously improve with the increasing data ratio, likely due to excessive tool call reducing the efficiency of the model's reasoning. Based on the above analysis, we set the data ratio used in the experiment to 13:7.

\section{Case Study}
\label{app:case study}
To better demonstrate the superiority of Tool-Light over other methods in efficiency and necessity metrics, we conduct a case study. We compare the reasoning trajectories of Tool-Light with those of Tool-Star. As shown in Table~\ref{tab:case_math500}, when addressing the same problem, even if the final answer is correct, Tool-Light's result achieves fewer tool calls compared to that of Tool-Star. And as shown in Table~\ref{tab:case_aime24}, when faced with a more difficult problem, the Tool-Light method will proactively call external tools to assist in answering questions when necessary.
\begin{table*}[!t]
    \centering
    \caption{A comparison example between Tool-Light and Tool-Star on MATH500 dataset.}
    \fontsize{9pt}{11pt}\selectfont
    \renewcommand{\arraystretch}{0.8} 
    \begin{tabular}{p{0.95\linewidth}}
    \midrule
        \rowcolor{gray!20}\textbf{Example \#1 from MATH500 Dataset} \\
    \midrule
        \textbf{Question:} \\
        The square root of $t$ is greater than $2$ and less than $3.5$. How many integer values of $t$ satisfy this condition?\\
        \textbf{Golden Answer:} \\
        8 \\
    \midrule
        \rowcolor{gray!20}
        \textbf{Tool-Light's TIR Stragety}\\
    \midrule
        $<$think$>$ To determine how many integer values of $t$ satisfy the condition $ 2 < \sqrt{t} < 3.5$, we can follow these steps:

1. Square all parts of the inequality to eliminate the square root:
   \[
   2^2 < t < (3.5)^2
   \]
   This simplifies to:
   \[
   4 < t < 12.25
   \]

2. Identify the integer values of \( t \) that fall within this range. The integers between 4 and 12.25 are 5, 6, 7, 8, 9, 10, 11, and 12.

3. Count the number of these integers.

Let's implement this in Python to ensure accuracy.
 $<$/think$>$$<$python$>$
 \begin{lstlisting}[style=python]
# Define the range
lower_bound = 4
upper_bound = 12.25

# Find the integer values in the range
integer_values = [t for t in range(int(lower_bound) + 1, int(upper_bound) + 1)]

# Count the number of integer values
num_integer_values = len(integer_values)
print(num_integer_values)
\end{lstlisting}
$<$/python$>$$<$result$>$
8
$<$/result$>$ $<$answer$>$The number of integer values of \( t \) that satisfy the condition \( 2 < \sqrt{t} < 3.5 \) is \(\boxed{8}\).$<$/answer$>$\\
    \midrule
        \rowcolor{gray!20}
        \textbf{Tool-Star's TIR Stragety}\\
    \midrule
    $<$think$>$Okay, so I need to figure out how many integer values of t satisfy the condition that the square root of $t$ is between 2 and 3.5. Hmm, let me start by understanding what exactly the question is asking. 

First, the square root of t has to be greater than 2 and less than 3.5. That means $\sqrt(t) > 2$ and $\sqrt(t) < 3.5$. But since we're dealing with real numbers, maybe even irrational ones, but the question specifies that t must be an integer. So $t$ has to be a perfect square between 4 (since 2 squared is 4) and... well, 3.5 squared. Let me compute that.

$<$/think$>$$<$python$>$
\begin{lstlisting}[style=python]
lower = 2 ** 2
upper = 3.5 ** 2
print(f'Lower bound: {lower}, Upper bound: {upper}')
\end{lstlisting}
$<$/python$>$$<$result$>$
Lower bound: 4, Upper bound: 12.25
$<$/result$>$So $\sqrt(t)$ needs to be strictly greater than 2, which gives $t > 4$, and strictly less than 3.5, which gives $t < (3.5)^2 \approx 12.25$. Therefore, t must be integers from 5 up to 12 inclusive. To confirm:

$<$/think$>$$<$python$>$
\begin{lstlisting}[style=python]
start = int(2**2) + 1
end = int(3.5**2)
count = end - start + 1
print(f'Start: {start}, End: {end}, Count: {count}')
\end{lstlisting}
$<$/python$>$$<$result$>$
Start: 5, End: 12, Count: 8
$<$/result$>$$<$think$>$Therefore, the number of integers satisfying the condition is indeed 8.
$<$/think$>$$<$answer$>$
\boxed{8}
$<$/answer$>$\\
    \bottomrule
    \end{tabular}
    \label{tab:case_math500}
\end{table*}

\begin{table*}[!t]
    
    \centering
    \caption{A comparison example between Tool-Light and Tool-Star on AIME24 dataset.}
    \fontsize{9pt}{11pt}\selectfont
    \renewcommand{\arraystretch}{0.8} 
    \begin{tabular}{p{0.95\linewidth}}
    \midrule
        \rowcolor{gray!20}\textbf{Example \#2 from AIME24 Dataset} \\
    \midrule
        \textbf{Question:} \\
        Find the largest possible real part of \[(75+117i)z+\frac{96+144i}{z}\]where $z$ is a complex number with $|z|=4$.\\
        \textbf{Golden Answer:} \\
        540 \\
    \midrule
        \rowcolor{gray!20}
        \textbf{Tool-Light's TIR Stragety}\\
    \midrule
        $<$think$>$ To solve the problem, we need to find the maximum value of the expression \((75 + 117i)z + \frac{96 + 144i}{z}\) where \(z\) is a complex number with \(|z| = 4\).

Let's represent \(z\) as \(4e^{i\theta}\), where \(\theta\) is a real number. Then we can write:
\[ (75 + 117i)z + \frac{96 + 144i}{z} = (75 + 117i)4e^{i\theta} + \frac{96 + 144i}{4e^{i\theta}}. \]

This can be simplified to:
\[ 4(75 + 117i)e^{i\theta} + \frac{24 + 36i}{e^{i\theta}}. \]

\textbf{(***More reasoning contentare omitted here***)}
\begin{lstlisting}[style=python]
import numpy as np

# Define the complex numbers
a = 75 + 117j
b = 96 + 144j

# Define the magnitude of z
r = 4

# Initialize the maximum real part
max_real_part = -np.inf
max_theta = 0

# Evaluate the expression for a range of theta values
for theta in np.linspace(0, 2 * np.pi, 1000):
    z = r * np.exp(1j * theta)
    expr = a * z + b / z
    real_part = np.real(expr)
    if real_part > max_real_part:
        max_real_part = real_part
        max_theta = theta

print(max_real_part)
\end{lstlisting}
$<$/python$>$$<$result$>$
539.9979693595761
$<$/result$>$$<$answer$>$The largest possible real part of the expression \((75 + 117i)z + \frac{96 + 144i}{z}\) where \(z\) is a complex number with \(|z| = 4\) is \(\boxed{540}\).$<$/answer$>$\\
    \midrule
        \rowcolor{gray!20}
        \textbf{Tool-Star's TIR Stragety}\\
    \midrule
$<$think$>$ To solve the problem, we need to find the maximum value of the expression \((75 + 117i)z + \frac{96 + 144i}{z}\) where \(z\) is a complex number with \(|z| = 4\). Let's represent \(z\) as \(4e^{i\theta}\), where \(\theta\) is a real number. This way, we can express \(z\) in terms of its polar form.

\textbf{(***More reasoning contentare omitted here***)}

However, we should confirm this by calculating the critical points. The derivative of \(324\cos\theta - 504\sin\theta\) is \(-324\sin\theta - 504\cos\theta\). Setting this to zero gives us the critical points. But since we know the maximum value of the expression \(a\cos\theta + b\sin\theta\) is \(\sqrt{a^2 + b^2}\), we can directly use this formula.

Thus, the largest possible real part is:

$<$/think$>$$<$answer$>$\boxed{594}
$<$/answer$>$\\
    \bottomrule
    \end{tabular}
    \label{tab:case_aime24}
\end{table*}

\end{document}